\documentclass[conference]{IEEEtran}

\DeclareRobustCommand{\bigO}{%
  \text{\usefont{OMS}{cmsy}{m}{n}O}%
}

\usepackage[pdftex]{graphicx}
\usepackage{amsmath}
\usepackage{amsfonts}
\usepackage{xcolor}
\usepackage{bbm}

\usepackage[caption=false,font=normalsize,labelfont=sf,textfont=sf]{subfig}
\DeclareMathOperator{\Tr}{Tr} 

\begin{document}
\bstctlcite{IEEEexample:BSTcontrol}
\title{Photonic co-processors in HPC: using LightOn OPUs for Randomized Numerical Linear Algebra}

\author{\IEEEauthorblockN{\hspace{5em}Daniel Hesslow,
    Alessandro Cappelli,
    Igor Carron, 
    Laurent Daudet,
    Raphaël Lafargue,
	\\
	\hspace{5em}
	Kilian Müller,
	Ruben Ohana, 
	Gustave Pariente, and 
	Iacopo Poli}
\IEEEauthorblockN{\textit{LightOn, Paris, France},\\
{\tt \{firstname\}@lighton.ai}} \vspace*{-8mm}
}

\maketitle

\begin{abstract}
Randomized Numerical Linear Algebra (RandNLA) is a powerful class of methods, widely used in High Performance Computing (HPC). RandNLA provides approximate solutions to linear algebra functions applied to large signals, at reduced computational costs. However, the randomization step for dimensionality reduction may itself become the computational bottleneck on traditional hardware. Leveraging near constant-time linear random projections delivered by LightOn Optical Processing Units we show that randomization can be significantly accelerated, at negligible precision loss, in a wide range of important RandNLA algorithms, such as RandSVD or trace estimators.
\end{abstract}

\IEEEpeerreviewmaketitle

\section{Introduction}
At extreme scales, the exact computation of classic operations such as matrix multiplication or Singular Value Decomposition (SVD) becomes remarkably difficult - with sophisticated distributed schemes - or even unfeasible. As an alternative, Randomized Numerical Linear Algebra (RandNLA) provides approximate results at reduced computational costs. In some instances, RandNLA methods are the only way to obtain results in tractable time. These methods are being increasingly used in HPC, as evidenced in the recent RASC workshop \cite{rasc_report}. 
The principle of RandNLA is to reduce the dimensionality of the objects being processed, so that linear algebra operations are performed in a compressed space, at reduced time and memory cost, while approximately preserving the result. However, this assumes that the dimensionality reduction step, based on randomization, is not itself a computational bottleneck.

The baseline method for randomized compression is using Gaussian random projections, i.e. the multiplication by a $n\times m$ random matrix whose entries are drawn from a Gaussian distribution. Gaussian random projections are often used for their ability to lead to provable theorems for approximation bounds, that are essential in many practical settings.
However, computing such projections in large dimensions is costly on  CPU / GPU hardware, as they involve the generation and processing of large arrays of random numbers, in $\bigO(n m)$ computational complexity for processing a single input vector. 

We present a radically different approach, with specialized photonic hardware for Gaussian random projections. The LightOn Optical Processing Unit (OPU) \cite{brossollet}, now commercially available and originally designed for AI applications, is a co-processor that leverages free-space photonics to compute Gaussian random projections \textit{in} $\bigO(1)$  \textit{near constant time} (currently at $\sim 1.2$  ms, with a  $\times 10-100$ speedup possible with the same technology), at dimensions scaling to up to 1 million in input and 2 millions in output. Delivering 1500 TeraOPS at 30 W \cite{power}, the OPU is typically two orders of magnitude more energy efficient for this operation than programmable silicon chips. In this study, we compare the performance of several RandNLA algorithms where the random projections are either performed on the OPU, or on high-end CPUs/GPUs. 


\newcommand{\norm}[1]{\left\lVert#1\right\rVert}




\section{RandNLA applications}

An OPU \cite{opu} natively performs the operation $r(x) = | Rx |^2$ where R is a fixed complex normal random matrix, whose coefficients are independent and identically distributed (i.i.d.), and $x$ is a binary input vector, with the element-wise squared modulus $| \cdot |^2$. Either optical or digital holography can be used to retrieve a real-valued linear random projection $g(x) = R x$. Using linear random projections, it is straightforward to handle multi-bit inputs, by successively processing bit-planes. 
Signed and floating point inputs can be handled in a similar manner.

\begin{figure*}[!t]
\centering
\subfloat[Matrix Multiplication]{\includegraphics[width=0.24\textwidth]{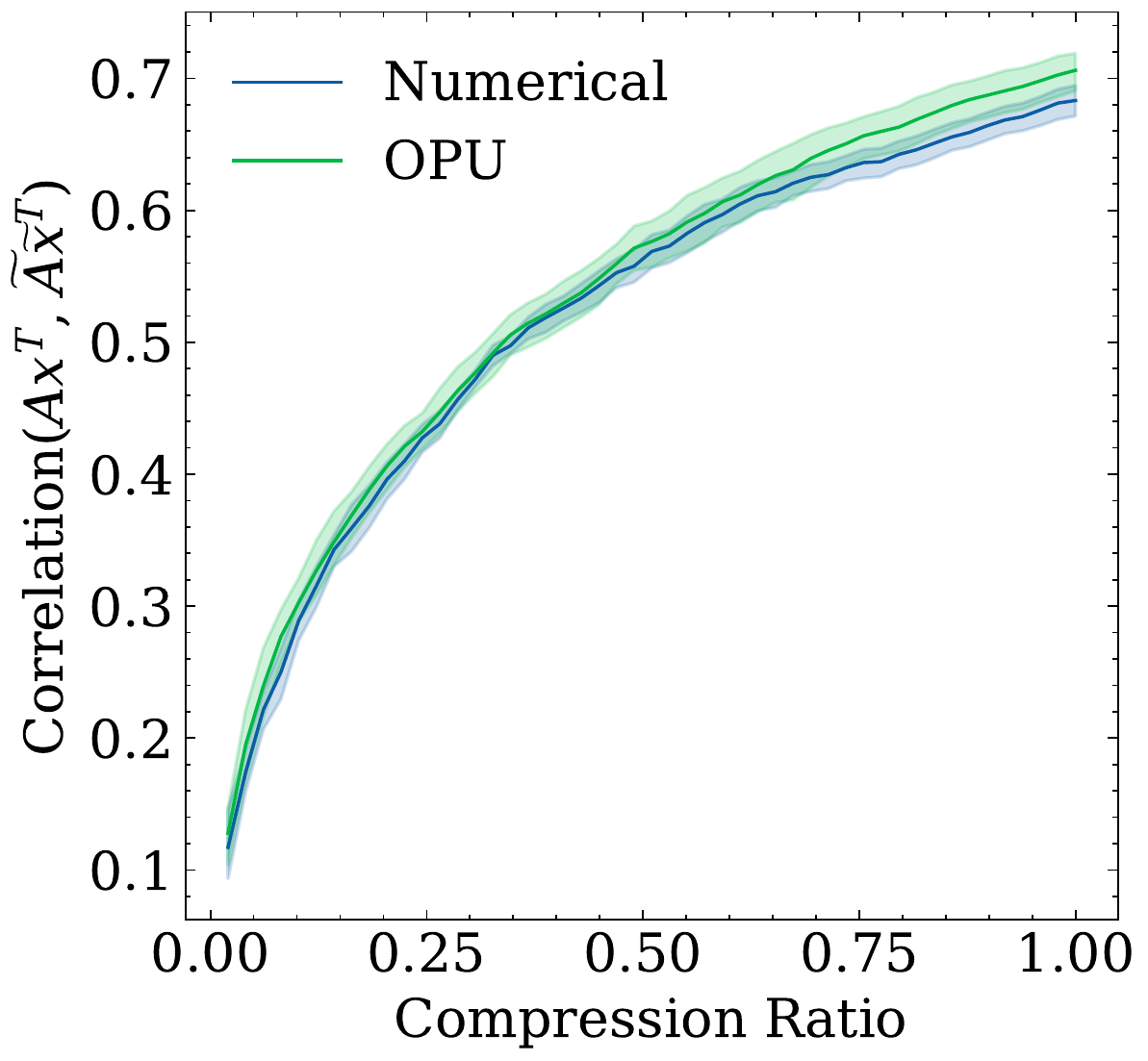}%
\label{fig_amm}}
\hfil
\subfloat[Trace Estimation]{\includegraphics[width=0.24\textwidth]{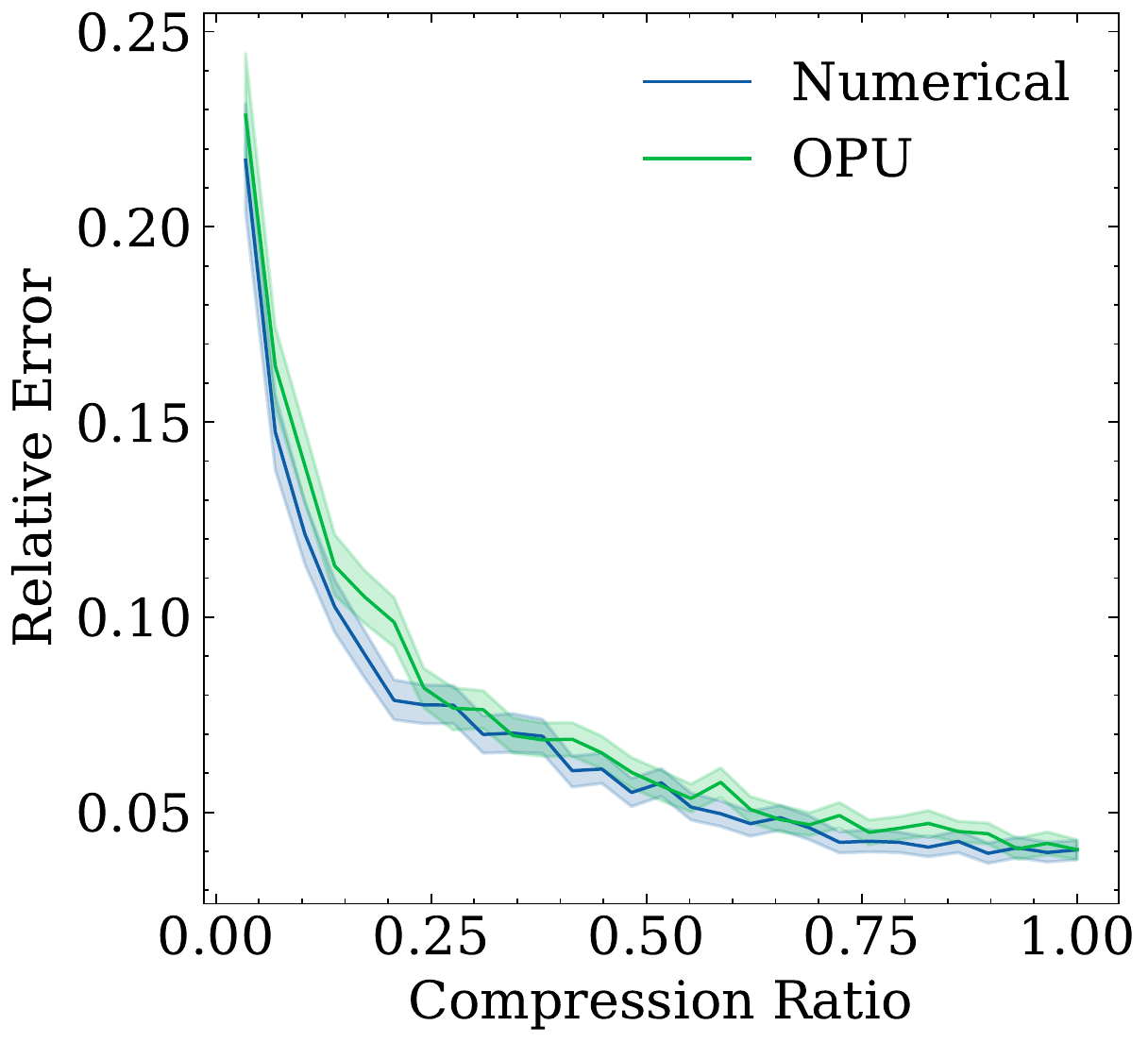}%
\label{fig_trace_est}}
\subfloat[Triangle Estimation]{\includegraphics[width=0.24\textwidth]{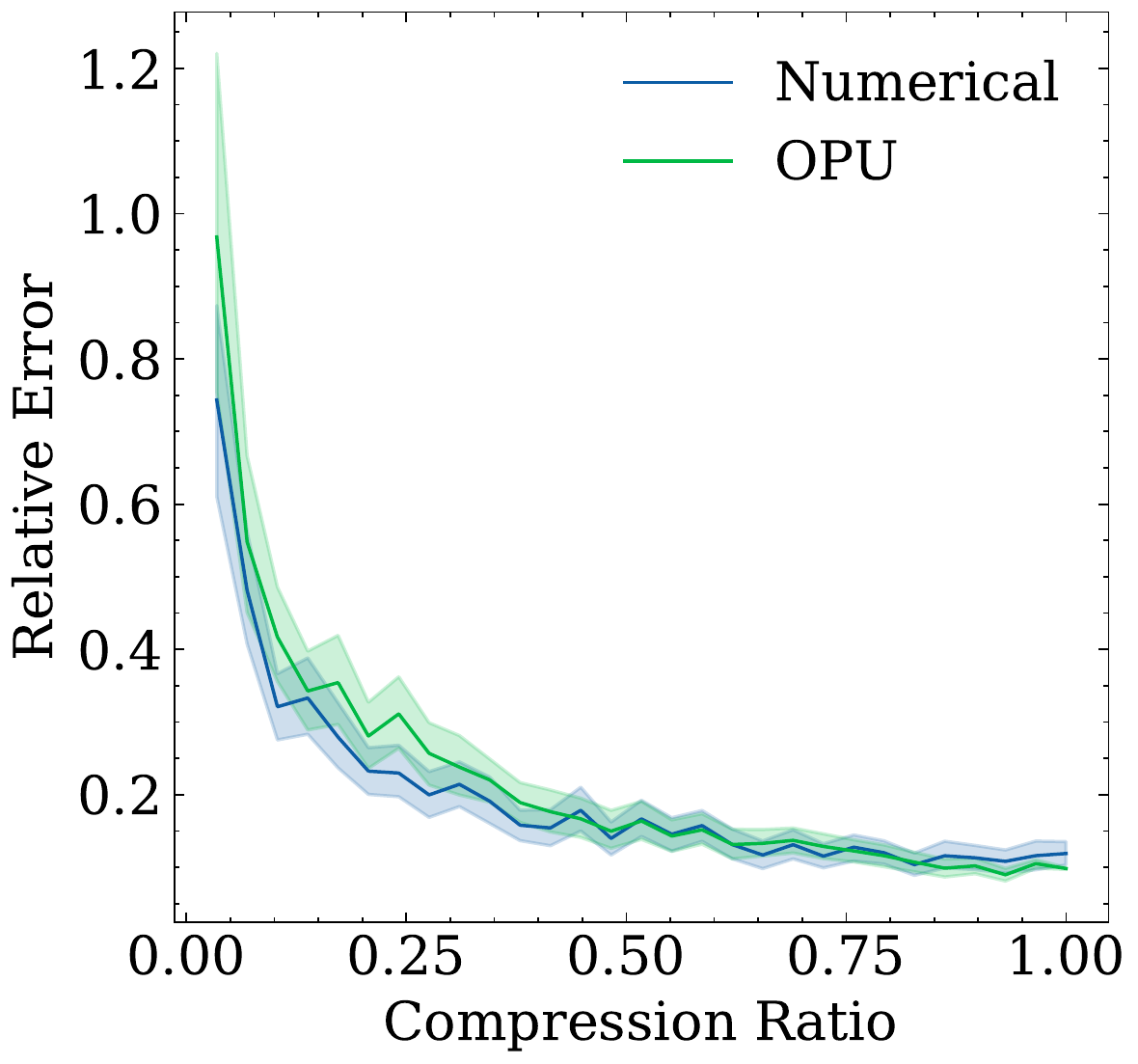}%
\label{fig_tri_est}}
\subfloat[rSVD]{\includegraphics[width=0.24\textwidth]{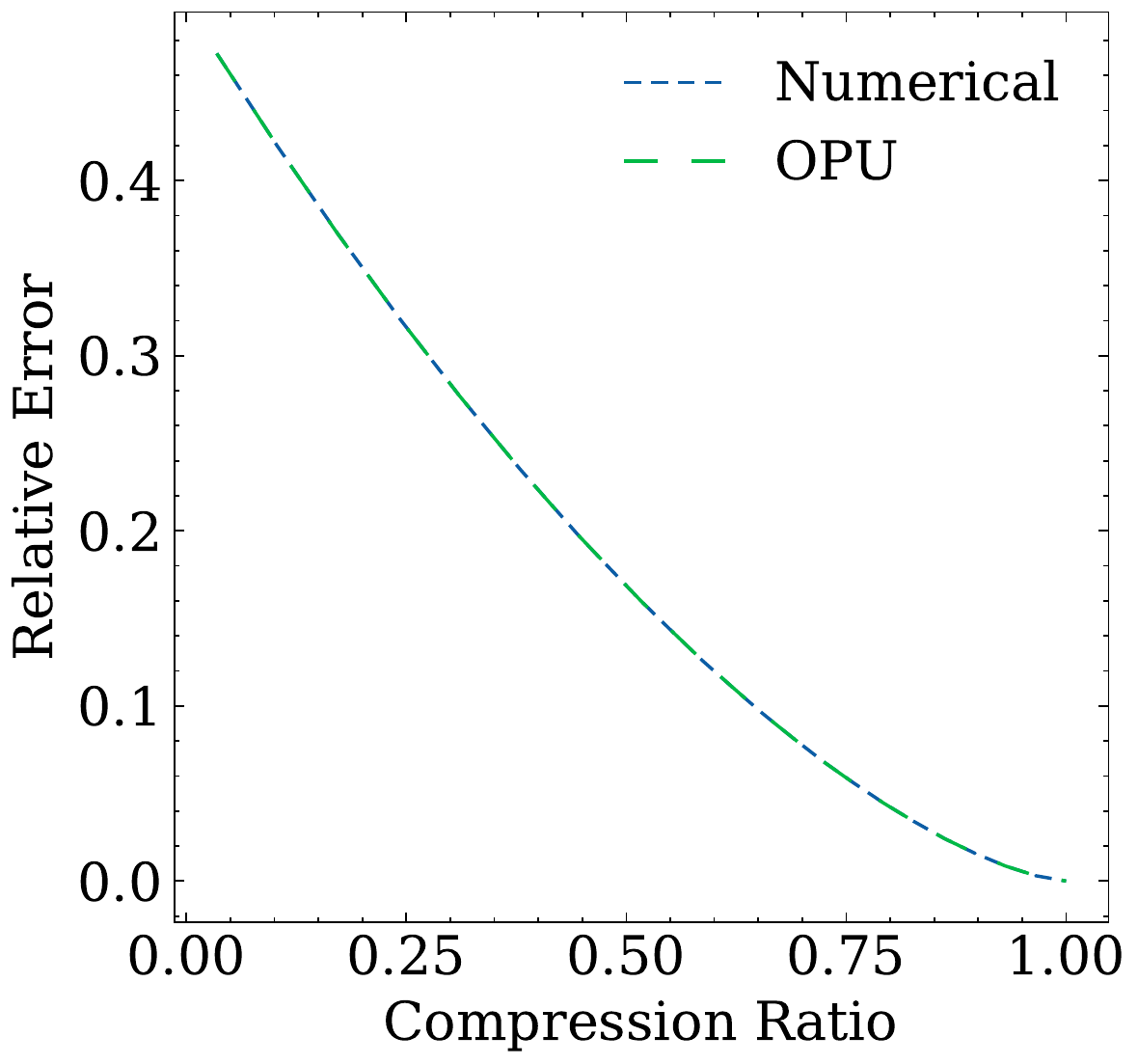}%
\label{fig_rsvd}}

\caption{Comparing the quality of the results of approximate matrix multiplication and trace estimation, triangle estimation and randomized SVD on the OPU with their numerical equivalents. We remark that the results obtained optically agree very well with the numerical results.}
\label{fig_sim}
\vspace*{-2mm}
\end{figure*}

\subsection{Approximate Matrix Multiplication}
While theoretical works have successively improved the computational complexity of matrix multiplication \cite{strassen, coppersmith}, the computational cost in practice still remains close to $\bigO(n^3)$. As a way to approximate matrix multiplication we will consider what is known as sketched matrix multiplication,
\begin{align}
    \widetilde{A} &=  R A, \hspace{1em} \widetilde{B} =  R B\\
    A^\top B &\approx A^\top R^\top R B =   \widetilde{A}^\top  \widetilde{B}
\end{align}
using the property that, for an i.i.d. random matrix $R$, $R^\top R \approx I$.
Let $R \in \mathbb{R}^{m \times n}$, and $A, B\in\mathbb{R}^{n\times n}$, the computational complexity of this operation is $\bigO(n^2m)$, which results in a $\tfrac{n}{m}$ speedup. We will refer to $\tfrac{m}{n}$  as the \textit{compression ratio}. 

\subsection{Trace Estimation}

Another fundamental operation in linear algebra is the computation of the trace of a matrix. 
\begin{align}
    \Tr(A) = \sum_{i=1}^n A_{ii} = \sum_{i=1}^n \lambda_i
\end{align}
where $\lambda_i$ are the eigenvalues.
While the computational complexity of the trace of a known matrix $A$ is $\bigO(n)$, there are many problems of the form $\Tr(f(A))$ where $f(A)$ is a potentially expensive matrix function. For such cases randomized techniques for trace estimation have been established. In particular we consider the \textit{Hutchinson’s estimator} \cite{hutchinson}, 
\begin{align}
    \Tr(A) \approx \Tr(RAR^\top)
\end{align}

An application of trace estimation is the computation of triangles in a graph, that is central in complex network analysis \cite{eubank}. 
This problem can be expressed as $\Tr(A^3)$ where $A \in \mathbb{R}^{n \times n}$ is the adjacency matrix of the graph. Combining sketched matrix multiplication with Hutchinson’s estimator:
\begin{align}
    \Tr(A^3) &\approx \Tr(R A^3 R^\top) \approx \Tr(RA AR^\top R AR^\top R A R^\top)\\
            &=  \Tr((R A R^\top)^3)
\end{align}

Its computational cost is $\bigO(m^3+n)$ using constant time random projections compared to the naive $\bigO(n^3)$.

\subsection{Randomized SVD}

Singular value decomposition (SVD) has numerous practical application throughout science and engineering.  However, computing the SVD becomes prohibitively expensive as the dimensions grow large - fundamentally at least $\bigO(m^2n)$. Randomized SVD (RandSVD) has drastically increased the size of the matrices that can be handled \cite{halko}. By finding a matrix $Q$ with $m$ orthogonal columns such that $A \approx  Q Q^\top A$, the SVD can be obtained by decomposing $Q^\top A$ instead of $A$, which can yield a considerable speedup. To obtain $Q$ one can find the closest orthogonal matrix to $AR$.
The full decomposition can be computed as:
\begin{align}
    U \Sigma V^T = \text{SVD}(Q^\top A), \hspace{1em} \text{SVD}(A) = (Q U) \Sigma V^T
\end{align}

\section{Performance}
We compare the OPU against an NVIDIA P100 GPU (16 GB RAM). For small random projections where input and output dimensions are smaller than $\sim 12 \times 10^3$ it is faster to perform the random projections on the GPU. After this point the OPU can bring large speedups. For very large random projections (with input / output sizes exceeding $7 \times 10^4$), which is fundamentally the target of RandNLA, the OPU is crucial as the GPU runs out of memory. Note that pre-/post-processing of the data brings a small linear $\bigO(n)$ overhead to the OPU.

\begin{figure}
    \centering
    \includegraphics[width = 0.44\textwidth]{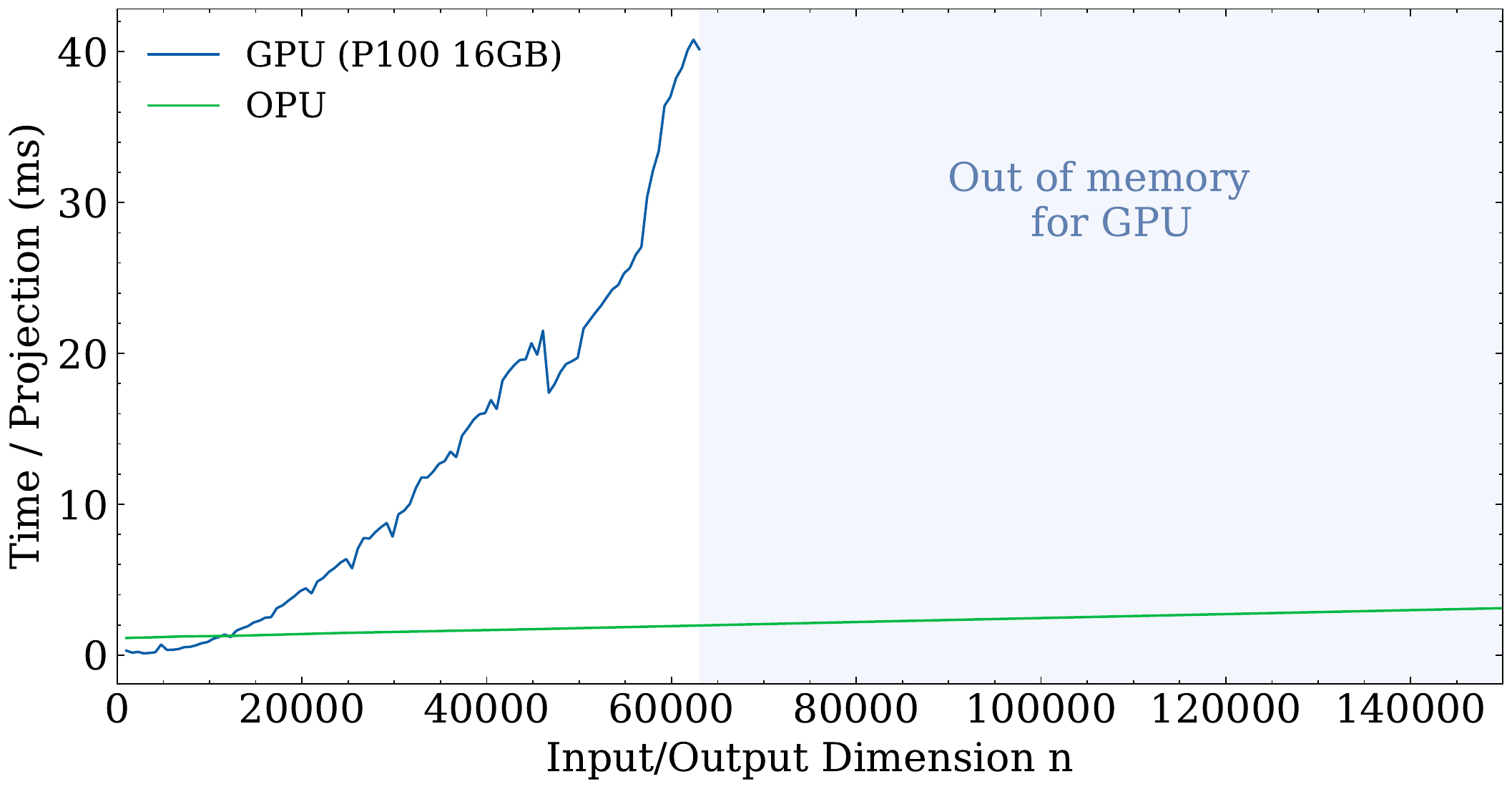}
    \caption{Comparison of performing a linear n by n random projection on the OPU and on a NVIDIA P100 16 GB GPU.}
    \label{fig:perf}
    \vspace*{-2mm}
\end{figure}

\section{Conclusion}
This study demonstrates that LightOn OPUs are well suited to hybrid pipelines for HPC relying on randomized linear algebra: the OPU performs the randomization step at high speed and low power consumption, while standard CPU/GPU hardware operates in the compressed domain. Experimental results show that the analog nature of the OPU photonic principle does not impact the end precision, compared to full precision digital randomization. At large data sizes, offloading an operation which on traditional hardware is $\bigO(n^2)$ onto specialized hardware producing the same computation in $\bigO(1)$ opens many directions both for research and practical engineering applications in HPC.

\bibliographystyle{IEEEtran}
\bibliography{IEEEabrv,main}
\clearpage
\onecolumn
\thispagestyle{empty}
\begin{center}
\includegraphics[width = 0.12\textwidth]{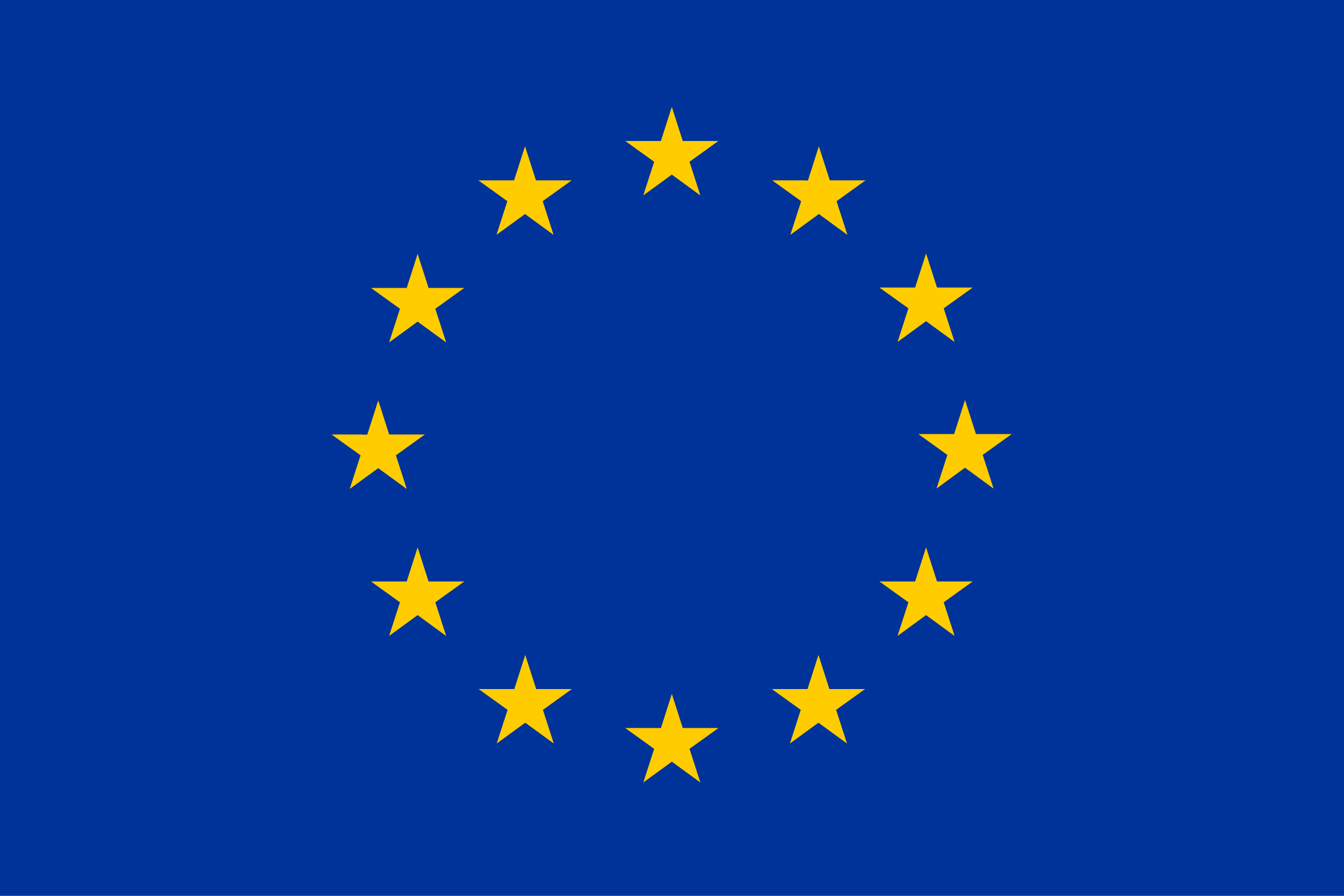}\\
This project has received funding from the European Union’s Horizon 2020 research and innovation programme under the Marie Skłodowska-Curie grant agreement No 860830
\end{center}
\end{document}